\documentclass[a4paper,fleqn]{cas-sc}

\usepackage[authoryear,longnamesfirst]{natbib}

\usepackage{amsmath}
\usepackage{amssymb}
\usepackage{mathtools}
\usepackage{amsthm}
\usepackage{amsfonts}
\usepackage{algorithmic}
\usepackage{algorithm}
\usepackage{multirow}
\usepackage{tikz}
\usepackage{pgfplots}
\usepgfplotslibrary{groupplots,dateplot}
\usetikzlibrary{positioning,decorations.pathreplacing,shadings,fit,backgrounds,patterns,shapes,arrows,arrows.meta,decorations.shapes,datavisualization}

\def\tsc#1{\csdef{#1}{\textsc{\lowercase{#1}}\xspace}}
\tsc{WGM}
\tsc{QE}

\makeatletter
\@fleqnfalse
\@mathmargin\@centering
\makeatother

\begin{document}
\let\WriteBookmarks\relax
\def\floatpagepagefraction{1}
\def\textpagefraction{.001}

\shorttitle{Leveraging Synergy for Imitation Learning}

\shortauthors{Y. Tanaka et~al.}

\title [mode = title]{SynIL: Leveraging Synergy for Offline Imitation Learning from Imperfect Demonstration Datasets}

\tnotemark[1]

\tnotetext[1]{This work was supported by JSPS KAKENHI Grant Number JP24K00841.}

%

\author[1]{Yuto Tanaka}
\ead{yuto.tanaka.r8@alumni.tohoku.ac.jp}

\credit{Conceptualization of this study, Methodology, Software, Analysis and interpretation of data, Drafting the article}

\affiliation[1]{organization={Department of Robotics, Graduate School of Engineering, Tohoku University},
            addressline={6-6-01 Aoba, Aramaki, Aoba-ku},
            city={Sendai},
            postcode={980-8579},
            state={Miyagi},
            country={Japan}}

\author[1]{Kyo Kutsuzawa}[orcid=0000-0002-5326-7847]
\ead{kutsuzawa@tohoku.ac.jp}

\credit{Conceptualization of this study, Methodology, Software, Analysis and interpretation of data, Drafting the article}

\author[1]{Martina Doku}[orcid=0009-0008-8022-9206]
\cormark[1]
\ead{martina.doku.t7@dc.tohoku.ac.jp}

\credit{Conceptualization of this study, Methodology, Drafting the article}

\author[1]{Dai Owaki}[orcid=0000-0003-1217-3892]
\ead{owaki@tohoku.ac.jp}

\credit{Conceptualization of this study, Methodology, Drafting the article}

\author[1]{Mitsuhiro Hayashibe}[orcid=0000-0001-6179-5706]
\cormark[1]
\ead{hayashibe@tohoku.ac.jp}
\ead[url]{https://neuro.mech.tohoku.ac.jp/}

\credit{Conceptualization of this study, Methodology, Drafting the article}

\cortext[1]{Corresponding author}


\begin{abstract}
Imitation learning enables robots to acquire complex skills directly from massive demonstration datasets, but its performance degrades severely when datasets are contaminated with suboptimal or noisy demonstrations. While prior quality-assessment methods attempt to filter or reweight data, they typically rely on manual pre-selection of expert reference data or task-specific heuristics, limiting scalability. To address this challenge, we introduce SynIL (Synergy-based Imitation Learning), a novel framework for automated, label-free demonstration quality assessment in offline reinforcement learning. Grounded in neuroscientific evidence that motor synergy—a low-dimensional coordinated structure in movement—correlates directly with motor proficiency, SynIL algorithmically quantifies synergy manifestation to generate dense, transition-level reward signals via self-supervised reward regression. Comprehensive evaluations on D4RL locomotion benchmarks and multi-human Robomimic manipulation datasets demonstrate that synergy-derived rewards correlate strongly with ground-truth rewards. Furthermore, SynIL substantially outperforms Behavior Cloning (BC) and achieves performance comparable to—and in sparse-reward human teleoperation scenarios, superior to—offline reinforcement learning trained on true environment rewards.
\end{abstract}


\begin{highlights}
\item Synergy, a coordinated modular structure proposed in neuroscience, enables automated, label-free demonstration quality assessment in mixed-proficiency datasets.
\item The proposed synergy scores provide dense transition-level rewards for offline reinforcement learning through self-supervised reward regression.
\item The proposed SynIL framework significantly outperforms Behavior Cloning on D4RL benchmarks and substantially improves task success rates on multi-human Robomimic environments.
\end{highlights}


\maketitle

\section{Introduction}

\begin{figure*}
    \centering
    \resizebox{\textwidth}{10cm}{
        \input{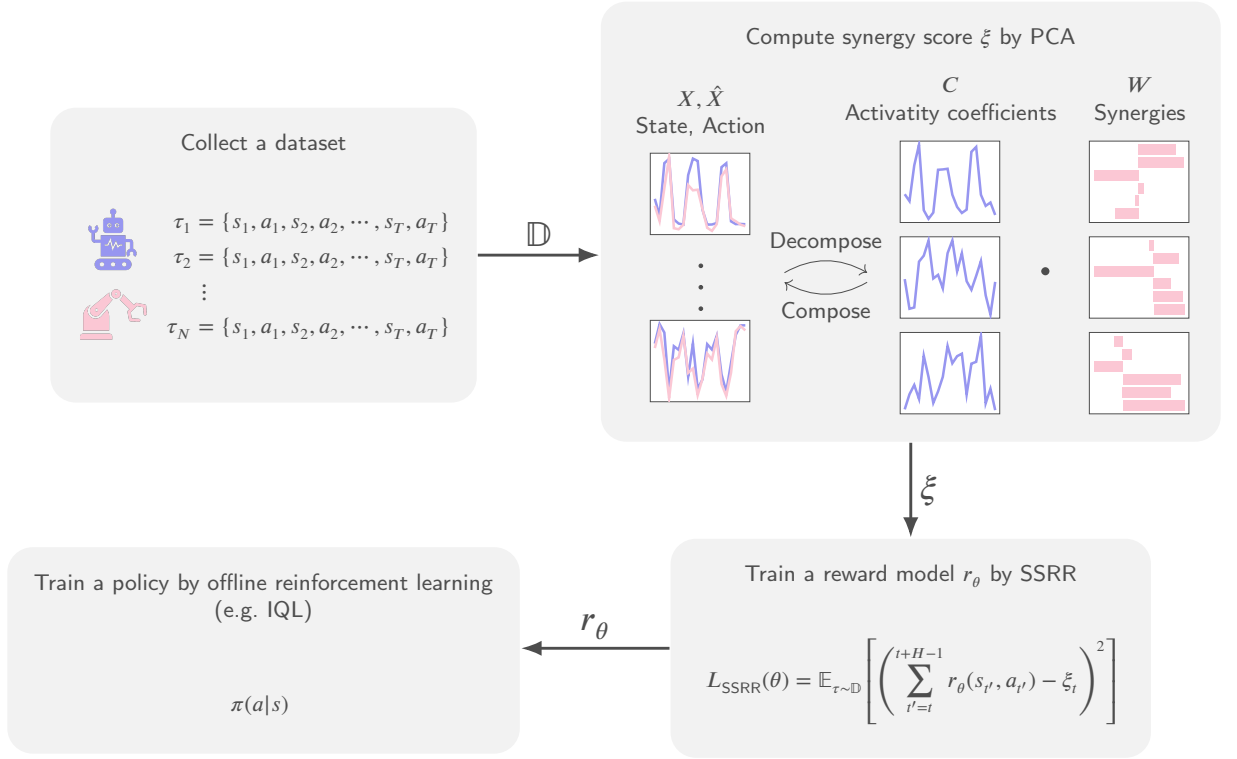}
    }
    \caption{Overview of SynIL.
        Synergy scores $\xi$ are calculated from trajectories of a dataset $\mathbb{D}$ (top-right), and a reward model is trained based on the synergy scores (bottom-right).
        Offline reinforcement learning is finally performed with rewards computed by the trained reward model (bottom-left).}
    \label{fig:diagram}
\end{figure*}

Imitation learning, a method for mimicking expert demonstrations, has gained attention as a new learning paradigm that can potentially outperform manual programming requiring considerable time and a high level of expertise in coding and control.
Unlike traditional manual programming methods, imitation learning is a data-driven approach, allowing learning without the need for complex design, programming skills, and specialized knowledge.
Imitation learning is highly regarded as a more realistic approach in fields such as robotics, healthcare, and autonomous driving \citep{Zare2024,kim2407surgical,pan2017agile}.
In robotics, imitation learning has successfully automated numerous actions across various robots, as demonstrated by robotic foundation models \citep{Kawaharazuka2025}.

In spite of its potential, the performance of imitation learning is adversely affected by low quality teaching data \citep{robomimic2021,Belkhale2023}.
A large amount of optimal demonstration data is required to ensure the stable performance of imitation learning.
However, due to factors such as the difficulty of the task, the skill level of the expert, and issues like fatigue and concentration when collecting demonstration data, the data obtained may include suboptimal data.
As the dataset grows larger, the likelihood of including incomplete data and noise increases.
Presence of suboptimal or poor-quality demonstrations gets imitation learning to mimic these poor-quality demonstrations, resulting in decreased performance.

Evaluating the quality of demonstrations will overcome the above issue.
For example, DemoDICE \citep{kim2022demodice}, ISWBC \citep{li2024imitation}, and DWBC \citep{xu2022discriminator} pre-divide the training data into optimal data $D_E$ from experts and $D_U$ with unknown proficiency, enhancing imitation learning performance by focusing more on expert data or adjusting the distributional distance with non-optimal data.
Additionally, CLUE \citep{liu2023clue} and ORIL \citep{zolna2020offline} address this issue by converting distributional distances with optimal data into rewards and conducting offline reinforcement learning, which considers these rewards.
This approach has the potential to enable to learn behaviors beyond the (possibly poor) quality of the demonstrations.
However, in all these studies, it is necessary for humans to pre-determine the expert data.
Furthermore, when it comes to qualitative evaluation, it may be challenging to discern which instructional actions are optimal depending on the task.
Therefore, a method to automatically and quantitatively evaluate optimality is needed.

In this study, we focus on the concept of synergy, which is proposed in neuroscience, for automated demonstration-quality assessment.
Synergy refers to the low-dimensional structure of coordinated muscles and joints, which are often observed in human proficient movements \citep{latash2014dexterity,turvey1990coordination}.
Many studies have reported that the degree of synergy manifestation corresponds to the proficiency of movements \citep{zaal1999unlearned,gentner2010encoding}.
This phenomenon has also been observed from reinforcement learning in locomotion \citep{chai2020motor} and reaching movements \citep{han2021synergy}.
Synergies and their degree of manifestation can be obtained algorithmically.
Thus, we expect that synergy can be used for automated demonstration-quality assessment.

In this study, we propose a method of imitation learning with automatic assessment of demonstration quality.
We propose a method to compute rewards from unlabelled demonstrations based on synergy and use them in offline reinforcement learning.
Our contributions are as follows:
\begin{itemize}
    \item We propose a method to algorithmically compute the quality of demonstrations based on synergy and use it as reward for learning.
    \item This study is the first application of the concept of synergy to assess the quality of demonstration movements in imitation learning.
    \item We successfully improved imitation learning performance across various datasets, including real human demonstrations.
\end{itemize}

\section{Related Work}

\subsection{Imitation Learning}

Imitation learning is a learning method aimed at mimicking the demonstrated actions of experts (such as humans) \citep{Zare2024}.
This approach enables robots to learn behaviors directly from expert behaviors, making it beneficial for tasks where manually programming movements or rules is difficult.
Standard imitation learning includes offline methods, which do not involve real environments, such as Behavior Cloning (BC).
In general, the policy $\pi_\theta$ with parameters $\theta$ learns to approximate the action $a$ taken in state $s$ to the state-action pairs $(s,a)$ in the expert's dataset $\mathcal{D}$.
The loss function for BC can be expressed as follows:
\begin{equation}
	L_\pi(\theta) = \mathbb{E}_{(s,a) \sim \mathcal{D}}[-\log\pi_\theta (a|s)]. \label{eqn:bc-loss}
\end{equation}
For continuous actions, a deterministic policy is often trained with a mean-squared error objective.
BC is fully offline and requires neither an explicit reward nor further interaction with the environment.
Recent robotic foundation models \citep{Kawaharazuka2025} are trained with almost the same procedure as BC.

Although BC is simple and powerful technique, it has several issues.
First, BC tends to accumulate approximation errors when the trained policy interacts with the environment in a closed loop, unlike the open-loop training procedure in Equation~\ref{eqn:bc-loss}.
It motivates extended techniques such as DAgger \citep{Ross2010}, inverse reinforcement learning \citep{Abbeel2004}, and adversarial learning \citep{Ho2016}.
Second, as Equation~\ref{eqn:bc-loss} assigns the same importance to all samples, its performance is sensitive to the quality and consistency of the demonstrations.
Offline reinforcement learning techniques will address this issue by associating rewards with states and actions.

\subsection{Offline Reinforcement Learning}

Standard reinforcement learning (online reinforcement learning) learns by interacting with the environment.
However, online reinforcement learning can be inefficient because it requires collecting data again to change the policy, and there is a risk of interacting with robots or vehicles online.
Therefore, while reinforcement learning research is conducted in simulation environments that consider safety and efficiency, there are still few applications in the real world.

Offline reinforcement learning is a method that learns only from a pre-collected static dataset without interacting with the environment.
Unlike online reinforcement learning, offline reinforcement learning does not interact with the real world, making it effective in settings where data collection is high-cost or dangerous, and online interaction is impractical (such as in robotics \citep{singh2022reinforcement} and autonomous driving \citep{fang2022offline}).
In principle, any off-policy algorithm (an online reinforcement learning algorithm that predicts policies using past experiences) can be applied in offline reinforcement learning.
However, in many offline settings, off-policy algorithms perform poorly.
This is because the value function overestimates actions not supported by the dataset (OOD, Out-of-Distribution), which is due to distribution shifts where the distribution of the dataset differs from the distribution of the learned policy \citep{prudencio2023survey}.
Many offline reinforcement learning algorithms attempt to address distribution shifts, with representative methods including BCQ \citep{fujimoto2019off} and IQL \citep{kostrikov2021offline}.

IQL trains a policy with the following loss function:
\begin{equation}
	L_\pi(\theta) = \mathbb{E}_{(s,a) \sim \mathcal{D}}[- \exp(\beta(Q_{\phi}(s, a) - V_{\psi}(s))) \log\pi_\theta (a|s)], \label{eqn:iql-loss}
\end{equation}
where $Q_{\phi}(s, a)$ and $V_{\psi}(s)$ are the action-value and value functions, trained from rewards.
This can be viewed as an extension of BC, weighting each state-action pair by its advantage.
SynIL uses IQL because its weighted policy extraction provides a natural way to reflect continuous estimates of demonstration quality in the learned policy.

\subsection{Algorithms for Demonstration Quality}

Research has been conducted to improve the performance of imitation learning using datasets of varying quality.
For example, DemoDICE by \citep{kim2022demodice}, ISWBC by \citep{li2024imitation}, and DWBC \citep{xu2022discriminator} pre-divide the training data into optimal data $D_E$ from experts and $D_U$ with unknown proficiency, enhancing imitation learning performance by focusing more on expert data or adjusting the distributional distance with non-optimal data.
Additionally, CLUE by \citet{liu2023clue} and ORIL by \citet{zolna2020offline} address this issue by converting distributional distances with optimal data into rewards and conducting offline reinforcement learning, which considers these rewards.
However, in these studies, it is necessary for humans to pre-determine the expert data.
Furthermore, when it comes to qualitative evaluation, it may be challenging to discern which instructional actions are optimal depending on the task.
Therefore, a method to automatically and quantitatively evaluate optimality is needed.

Several studies have aimed to evaluate demonstration quality automatically.
\citet{zhang2026scizor} proposed a method for eliminating suboptimal samples from demonstrations automatically by using a self-supervised task progress predictor and a deduplication module.
\citet{Kulkarni2026} proposed using trajectory smoothness as a measure of demonstration quality to reduce the effect of noise and demonstrated that extracting smooth trajectories resulted in better performance than simply using the entire demonstrations.
However, these methods simply remove bad trajectories from training.
Even bad trajectories contain information about state transitions in the environment, so that information is also lost.
Also, \citet{Kulkarni2026} evaluated quality based on trajectories rather than state-action pairs.
This will eliminate the good parts of bad trajectories as well, reducing sample efficiency.
Therefore, scoring demonstration quality for each state-action pair or transition without eliminating samples would be effective.

\section{Methodology}

\subsection{Overview}

This section describes the proposed method, SynIL, which derives its name from synergy-based imitation learning.
This method allows a policy to imitate demonstrations by performing offline reinforcement learning based on rewards computed from the degree of synergy manifestation.
In SynIL, demonstrations do not have to include rewards; instead, rewards are computed from demonstrations using synergy, a concept proposed in neuroscience that quantifies the degree of movement coordination.
Unlike most conventional methods, SynIL does not require manual evaluation of demonstrations, except for setting several hyperparameters.

An overview of SynIL is provided in Figure~\ref{fig:diagram}.
First, the degree of synergy manifestation is quantified as a synergy score $\xi$.
Next, the synergy scores $\xi$ are converted into rewards by using self-supervised reward regression (SSRR) \citep{chen2021learning}.
Finally, offline reinforcement learning is performed using the synergy-based rewards associated with state-action pairs.

We assume a dataset
\begin{equation}
D = \left\{ \tau_i \right\}_{i=1}^{M}, \qquad \tau_i = \left\{ (\boldsymbol{s}_t, \boldsymbol{a}_t, \boldsymbol{s}_{t+1}) \right\}_{t=0}^{T_i-1}, \label{eq:dataset}
\end{equation}
where $\boldsymbol{s}_t$ and $\boldsymbol{a}_t$ indicate the state and action at time $t$, respectively.
No reward, success, or proficiency label is available during SynIL training.

\subsection{Acquisition of Synergy Scores based on Synergy Manifestation}

As the first step of SynIL, the quality of demonstration is quantified based on synergy.
Among various types of synergy, we employ the spatial synergy, which represents low-dimensional coordination patterns in multi-dimensional spatial data.
The spatial synergy is defined by the following equation:
\begin{equation}
    \boldsymbol{X}_t = \boldsymbol{W}_t^n \boldsymbol{C}_t^n + \bar{\boldsymbol{X}}_t + \mathrm{residuals} \label{eq.ssmt},
\end{equation}
where
\begin{equation}
    \boldsymbol{X}_t =
    \begin{bmatrix}
        \boldsymbol{s}_t & \boldsymbol{s}_{t+1} & \cdots & \boldsymbol{s}_{t+(H-1)} \\
        \boldsymbol{a}_t & \boldsymbol{a}_{t+1} & \cdots & \boldsymbol{a}_{t+(H-1)}
    \end{bmatrix}
    \label{eqn:xmat}.
\end{equation}
Here, data $\boldsymbol{X}_t$ is the $H$-step trajectory of the state-actions cut from a demonstraion, $n$ is the number of spatial synergies, $\boldsymbol{W}_t^n$ corresponds to the spatial synergies, and $\boldsymbol{C}_t^n$ represents the activation level of the spatial synergies at each time.
Also, $\bar{\boldsymbol{X}}_t$ incidates the row-wise mean values of $\boldsymbol{X}_t$.
Equation~\ref{eq.ssmt} is computed by the principle component analysys (PCA), which computes $\boldsymbol{W}_t^n$ and $\boldsymbol{C}_t^n$ to minimize $\mathrm{residuals}$.
It can be said that the smaller the $\mathrm{residuals}$ are, the more significant the spatial synergies are manifest.

Note that each state and action dimension is normalized using statistics from the entire training dataset before applying PCA.
This prevents PCA from letting variables with large numerical scales dominate the decomposition independently of their importance.
Finally, PCA can score how compactly the local movement segment can be represented, regardless of the global semantics of the entire trajectory and dataset.

Based on Equation~\ref{eq.ssmt}, the degree of synergy manifestation is quantified by a metric denoted by $(R^2)_t^n$, as follows:
\begin{equation}
    (R^2)_t^n = 1 - \frac{\| \boldsymbol{X}_t - \bar{\boldsymbol{X}}_t - \boldsymbol{W}_t^n \boldsymbol{C}_t^n \|_F^2}{\| \boldsymbol{X}_t - \bar{\boldsymbol{X}}_t \|_F^2} .
\end{equation}
Here, $\|\bullet\|_F$ represents the Frobenius norm.
$(R^2)_t^n$ ranges from $0$ to $1$, and the larger the value of $(R^2)$ is, the higher the degree of synergy manifestation is.

However, the above $(R^2)_t^n$ metric varies depending on the number of synergies, $n$.
Instead, this study uses the synergy level \citep{chai2020motor}, $\zeta_t$, as an indicator of synergy manifestation independent of $n$, defined as follows:
\begin{equation}
    \zeta_t = \frac{1}{N} \sum_{n = 1}^N (R^2)_t^n \label{eq.synergy_level},
\end{equation}
where $N$ indicates the degree of freedom of demonstrations.
Note that $(R^2)_t^0 \equiv 0$, and $(R^2)_t^n \equiv 1$ for $n \geq N$, so the synergy score sums up with a range of $n = 1, \ldots, N$.

In addition to the synergy level, $\zeta$, we also consider the standard deviation of the demonstrations.
This is because only use of the synergy level can lead to the problem of high evaluation of demonstrations with overly simplistic states and actions, such as a demonstration with no movement.
It is undesirable in many motor tasks to highly valuing the movement of doing nothing.
To address this issue, we use the following synergy score, $\xi_t$, that considers the standard deviation of state-action pairs as well as the synergy level:
\begin{equation}
    \xi_t = \log \zeta_t + \eta \log \bar{\sigma}_t \label{eq.synergy_score}.
\end{equation}
Here, $\bar{\sigma}_t$ is the mean value of the standard deviation of columns of $\boldsymbol{X}_t$.
$\eta$ is a hyperparameter to weight the standard deviation of the demonstration action.

\subsection{Reward Estimation Based on Synergy} \label{sec:synergy_reward_estimation}

As shown in Equations~\ref{eqn:xmat}--\ref{eq.synergy_score}, the synergy score $\xi_t$ are computed from an $H$-step state-action trajectory.
However, to use offline reinforcement learning, we need rewards associated with each state-action pair.
To bridge this gap, we consider that the synergy score $\xi_t$, computed from an $H$-step state-action trajectory, approximates the total reward of the $H$-step state-action trajectory, as expressed in the following equation:
\begin{equation}
    \xi_t \approx \sum_{t' = t}^{t + H - 1} r_{t'} \label{eqn:relationship between synergy score and rewards},
\end{equation}
where $r_{t'}$ indicates a reward at time $t'$.

To acquire such rewards, we construct a reward model that maps a state-action pair to a reward value.
Let a reward model be $r_{\theta}(\boldsymbol{s}_t, \boldsymbol{a}_t)$, where $\boldsymbol{s}_t$ and $\boldsymbol{a}_t$ denote a state and action at time $t$, respectively, and $\theta$ indicates the parameters of the reward model.
The reward model is trained to minimize the following loss function:
\begin{equation}
    L_{\text{SSRR}}(\theta) = \mathbb{E}_{\tau \sim \mathbb{D}} \left[ \left( \sum_{t' = t}^{t + H - 1} r_{\theta}(s_{t'}, a_{t'}) - \xi_t \right)^2 \right] \label{eqn:ssrr}.
\end{equation}
Minimizing this loss function gets the reward model to generate rewards which satisfy Equation~\ref{eqn:relationship between synergy score and rewards}.
This method is similar to the self-supervised reward regression (SSRR) \citep{chen2021learning}, with the noise level replaced with the synergy score.

\section{Evaluation}

\subsection{Dataset}

To evaluate the proposed method, we used D4RL \citep{fu2020d4rl} and Robomimic \citep{robomimic2021} datasets as demonstration datasets.
D4RL is a dataset for imitation learning and offline reinforcement learning, offering datasets for various environments and problem settings.
Among various tasks, we specifically used the locomotion dataset as a typical motor task.
The locomotion dataset includes three types of robots: Hopper, HalfCheetah, and Walker2d.
D4RL contains several types of datasets with varying quality in terms of the performance and rewards of the expert policies.
The ``medium'' dataset contains demonstrations with moderate quality, the ``medium-replay'' dataset is a mixed dataset of low to moderate quality demonstrations, and the ``medium-expert'' dataset is also a mix dataset of medium to high quality demonstrations.
Each dataset was generated by policies at different training stages of the reinforcement learning algorithm, soft actor-critic \citep{haarnoja2018soft}.
These datasets are suitable for our problem setting, as they involve varying-quality demonstrations.
It is known that typical imitation learning, such as behavior cloning (BC), decreases the performance with these types of datasets.

Robomimic, on the other hand, is an imitation-learning dataset with situations similar to real-world applications.
The dataset is composed of demonstraions in various tasks with a seven degrees-of-freedom manipulator with a gripper.
The demonstrations have been acquired by human operators with tele-operation in a simulator.
Among datasets in Robomimic, we used the multi-human (MH) datasets of Can and Square tasks.
The MH dataset contains a total of 300 demonstrations, with 50 collected from each of six individuals with varying levels of expertise.

\subsection{Problem Setting}

In the evaluation, we assume a dataset can be expressed as follows:
\begin{equation}
    \mathbb{D} = \left\{ (\boldsymbol{s}_t, \boldsymbol{a}_t, \boldsymbol{s}_{t+1}) \middle| t = 0, \ldots, T-1 \right\} \label{eq.dataset definition},
\end{equation}
where $\boldsymbol{s}_t$ and $\boldsymbol{a}_t$ indicates a state and action at time $t$.
We assume that the dataset does not have rewards or any sort of quality values for each tuple of $(\boldsymbol{s}_t, \boldsymbol{a}_t, \boldsymbol{s}_{t+1})$.
Although the D4RL datasets contains rewards, we excluded them to emulate the sitations where only demonstrations were provided.
Thus, reinforcement learning methods cannot be applied unless we compute reward values.

We compared the proposed method with BC, a common imitation learning method.
Additionally, for reference, we also compared them with the implicit Q-learning (IQL) \citep{kostrikov2021offline} with datasets containing the true rewards.
This comparison was conducted to discuss how our method, which uses estimated rewards, is comparable with offline reinforcement learning with true rewards.
The algorithm for SynIL is presented in Algorithm~\ref{alg:synil}.

\begin{algorithm}[tb]
    \caption{SynIL}
    \label{alg:synil}
    \begin{algorithmic}
        \REQUIRE dataset $\mathbb{D}$ of state-action trajectories, as defined in Equation~\ref{eq.dataset definition}
        \ENSURE a policy $\pi$ and a reward model $r_\theta$
        \FOR{each trajectory $\tau \sim \mathbb{D}$}
            \STATE Compute $\zeta_t$ and $\bar{\sigma}_t$ from $H$-step trajectories cut from $\tau$ by Equation=\ref{eq.synergy_level}
            \STATE Compute $\xi_t$ by Equation~\ref{eq.synergy_score}
        \ENDFOR
        \STATE Initialize reward function parameters $\theta$
        \FOR{each epoch}
            \FOR{each trajectory $\tau$ in $\mathbb{D}$}
                \STATE Compute loss $\mathcal{L}_{\text{SSRR}}(\theta)$ by Equation~\ref{eqn:ssrr}
                \STATE Update $\theta$ to minimize $\mathcal{L}_{\text{SSRR}}(\theta)$ through gradient descent
            \ENDFOR
        \ENDFOR
        \STATE Train a policy $\pi$ by offline reinforcement learning with the trained reward model $r_\theta$
    \end{algorithmic}
\end{algorithm}

\subsection{Hyperparameters}

The neural network and Synergy Score hyperparameters used in the SSRR are listed in Table~\ref{tab:hyperparameters_ssrr}.
The input dimension of the reward estimation model (RewardNet) was equal to the sum of the state and action dimensions, and the output dimension was one-dimensional since it represents the reward.
$H$ and $S$ indicate the horizon and stride; $H$-step trajectories are cut from a demonstration while shifting the start position by $S$-steps.
Additionally, $\eta$ was set through hand-tuning to achieve the highest correlation with the true reward.

Hyperparameters for SynIL, IQL, and BC are shown in Table~\ref{tab:hyperparameters_il}.
IQL becomes equivalent to the BC algorithm by setting the hyperparameter $\beta$ to 0.
Therefore, in D4RL, BC experiments were conducted by setting IQL's $\beta$ to 0.
The hyperparameters were hand-tuned to achieve the highest performance for each algorithm.

\begin{table}
    \centering
    \caption{Hyperparameters for SSRR}
    \begin{tabular}{llcc}
        \toprule
        \quad                          & Hyperparameter	         & Value in D4RL      & Value in Robomimic \\
        \midrule
        \multirow{3}{*}{RewardNet}     & Hidden layers           & 2                  & 2                  \\
                                       & Hidden Size             & 512                & 1024               \\
                                       & Activation              & \textit{ReLU}      & \textit{ReLU}      \\
                                       & Last Activation         & \textit{Linear}    & \textit{Linear}    \\
                                       & Epochs                  & 500                & 10000              \\
                                       & Batch Size              & 64                 & 64                 \\
                                       & Learning Rate           & $1 \times 10^{-4}$ & $1 \times 10^{-3}$ \\
                                       & Weight Decay            & $1 \times 10^{-2}$ & $ 0 $              \\
                                       & Criterion               & \textit{MSELoss}   & \textit{MSELoss}   \\
                                       & Optimizer               & \textit{Adam}      & \textit{Adam}      \\
        \midrule
        \multirow{3}{*}{Synergy Score} & Horizon $H$             & 200                & 20                 \\
                                       & Stride $S$              & 100                & 20                 \\
                                       & $\eta$                  & 0.05               & 0.05               \\
        \bottomrule
    \end{tabular}
    \label{tab:hyperparameters_ssrr}
\end{table}

\begin{table}
    \centering
    \caption{Hyperparameters for imitation learning}
    \begin{tabular}{lcc}
        \toprule
        Hyperparameter           & SynIL and IQL    & BC               \\
        \midrule
        Hidden size              & 256              & 256              \\
        Hidden layers            & 2                & 2                \\
        Activation               & ReLU             & ReLU             \\
        Optimizer                & Adam             & Adam             \\
        Epochs                   & $10^6$           & $10^6$           \\
        Evaluate period          & 5000             & 5000             \\
        Batch size               & 256              & 256              \\
        Learning rate            & $3\times10^{-4}$ & $3\times10^{-4}$ \\
        Discount                 & 0.99             & -                \\
        $\alpha$                 & 0.005            & -                \\
        $\tau$                   & 0.7              & -                \\
        $\beta$                  & 3.0              & -                \\
        \bottomrule
    \end{tabular}
    \label{tab:hyperparameters_il}
\end{table}

\subsection{Results}

\begin{figure*}[tb]
    \centering
    \begin{minipage}[b]{0.3\textwidth}
            \centering
            \includegraphics[width=\textwidth]{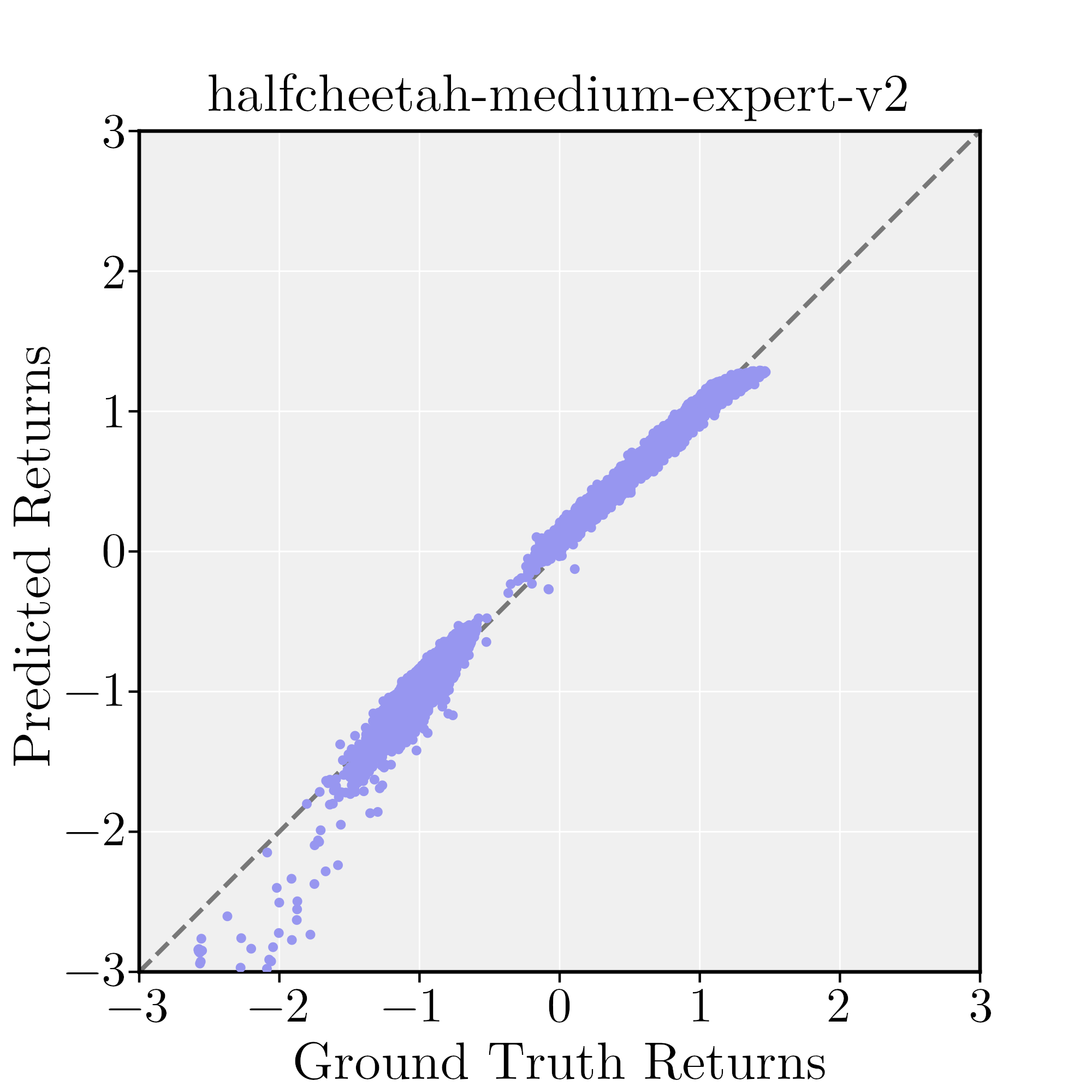}
    \end{minipage}
    \begin{minipage}[b]{0.3\textwidth}
            \centering
            \includegraphics[width=\textwidth]{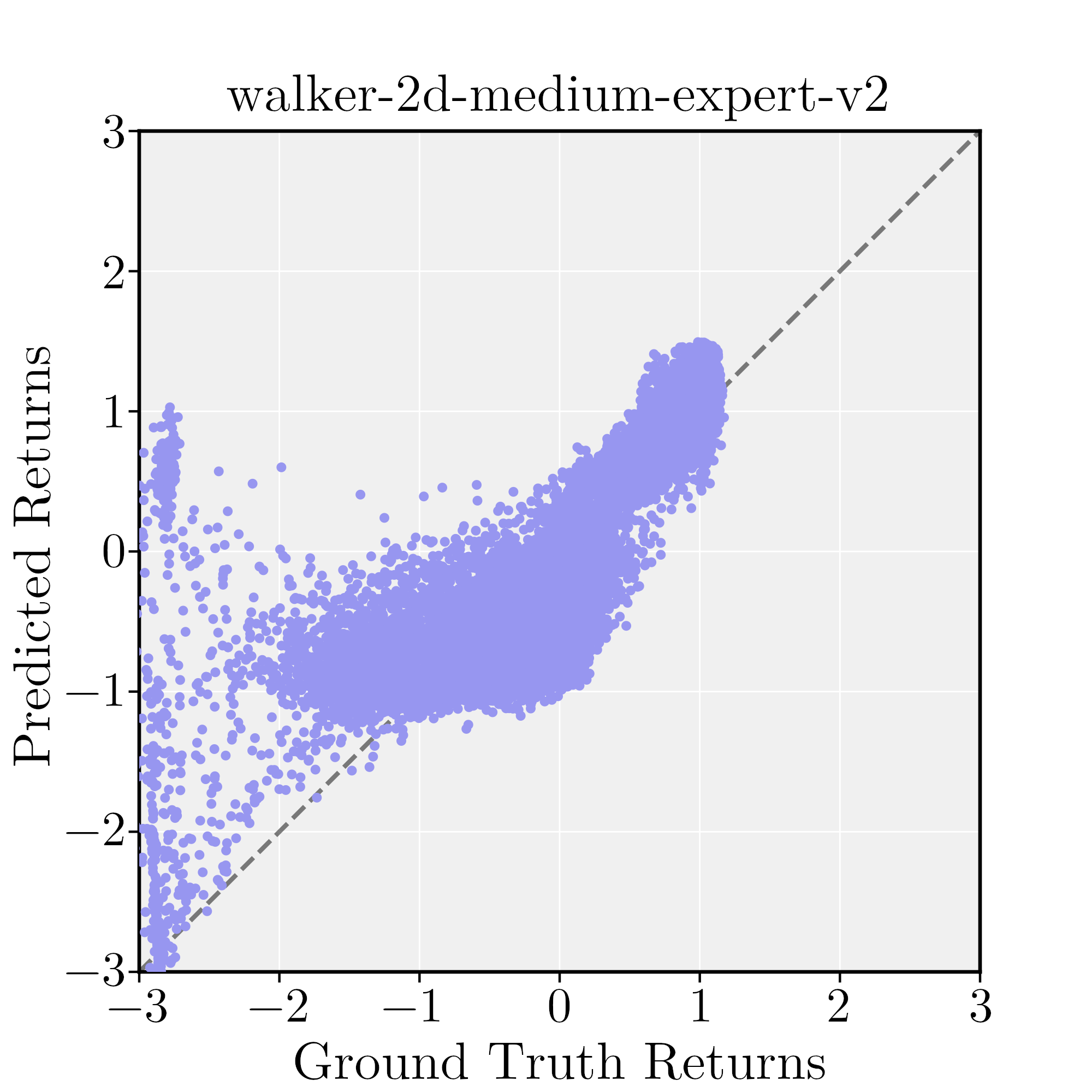}
    \end{minipage}
    \begin{minipage}[b]{0.3\textwidth}
            \centering
            \includegraphics[width=\textwidth]{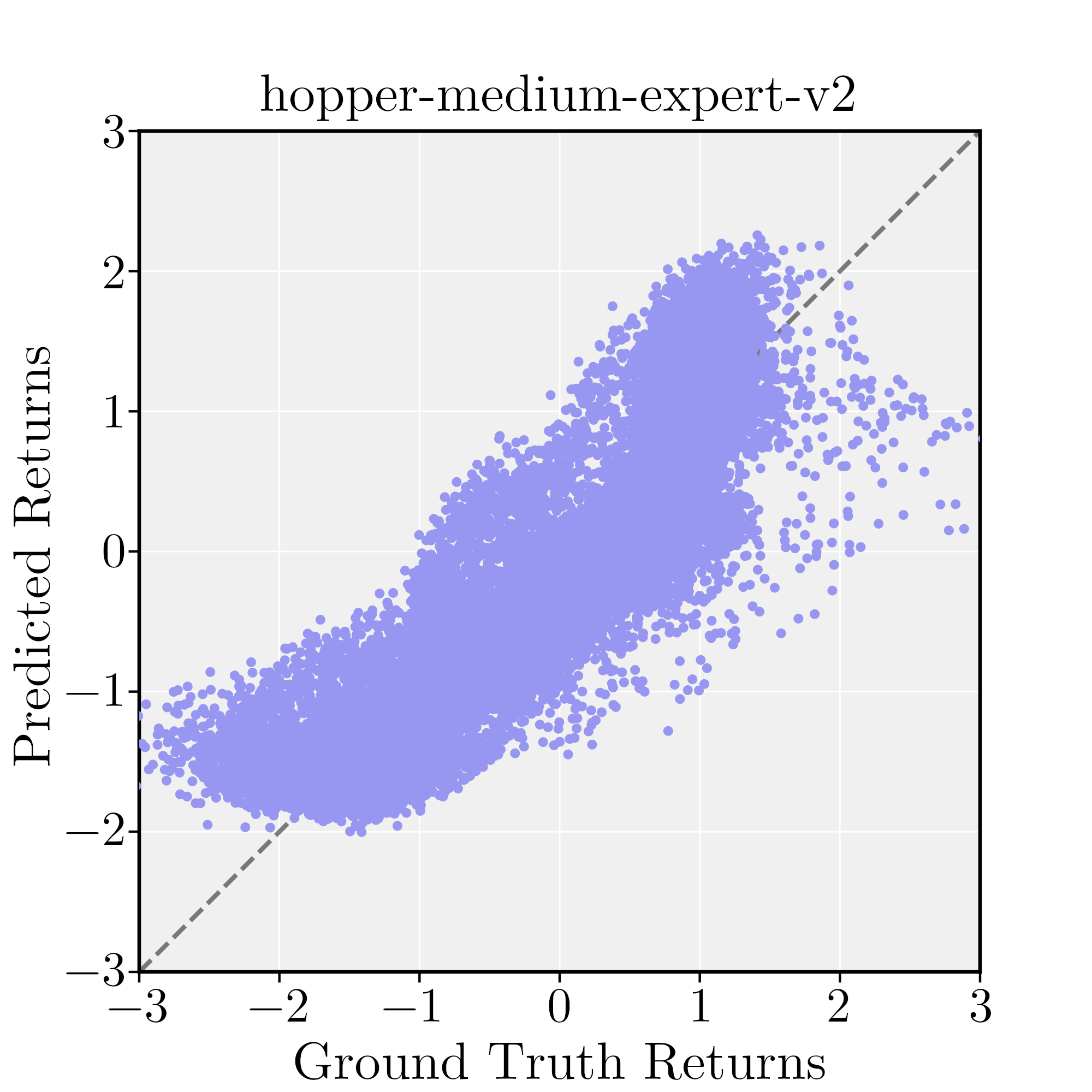}
    \end{minipage}
    \begin{minipage}[b]{0.3\textwidth}
            \centering
            \includegraphics[width=\textwidth]{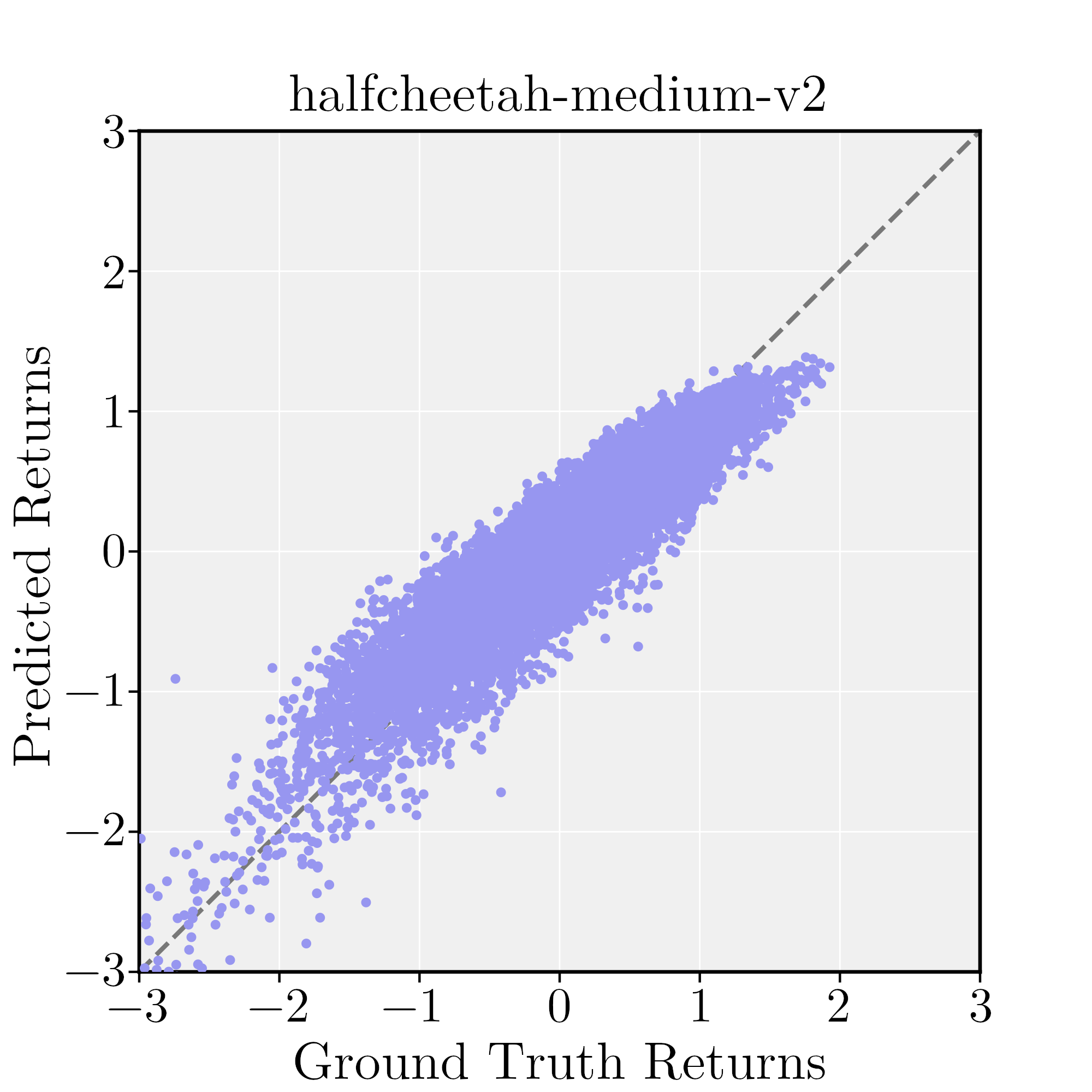}
    \end{minipage}
    \begin{minipage}[b]{0.3\textwidth}
            \centering
            \includegraphics[width=\textwidth]{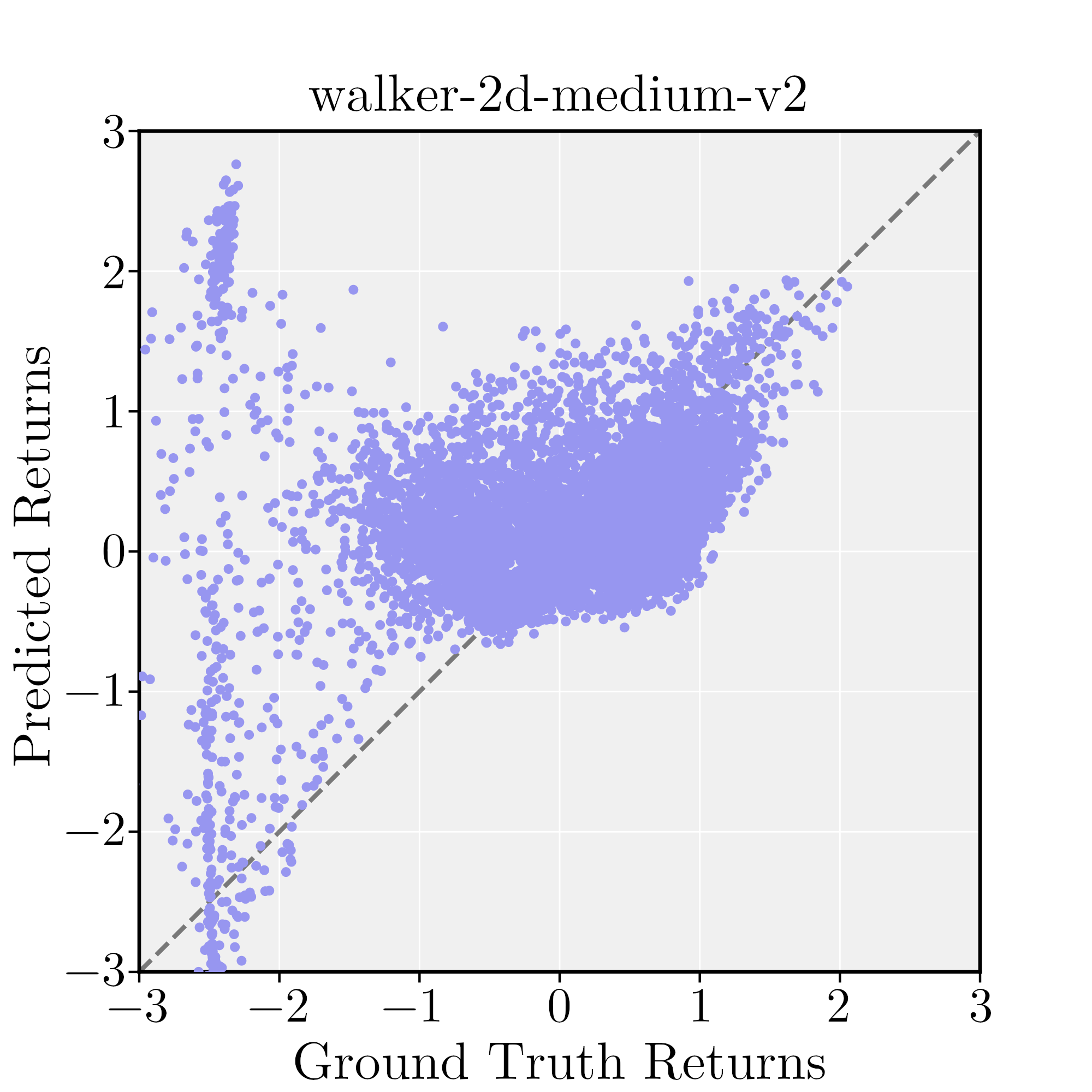}
    \end{minipage}
    \begin{minipage}[b]{0.3\textwidth}
            \centering
            \includegraphics[width=\textwidth]{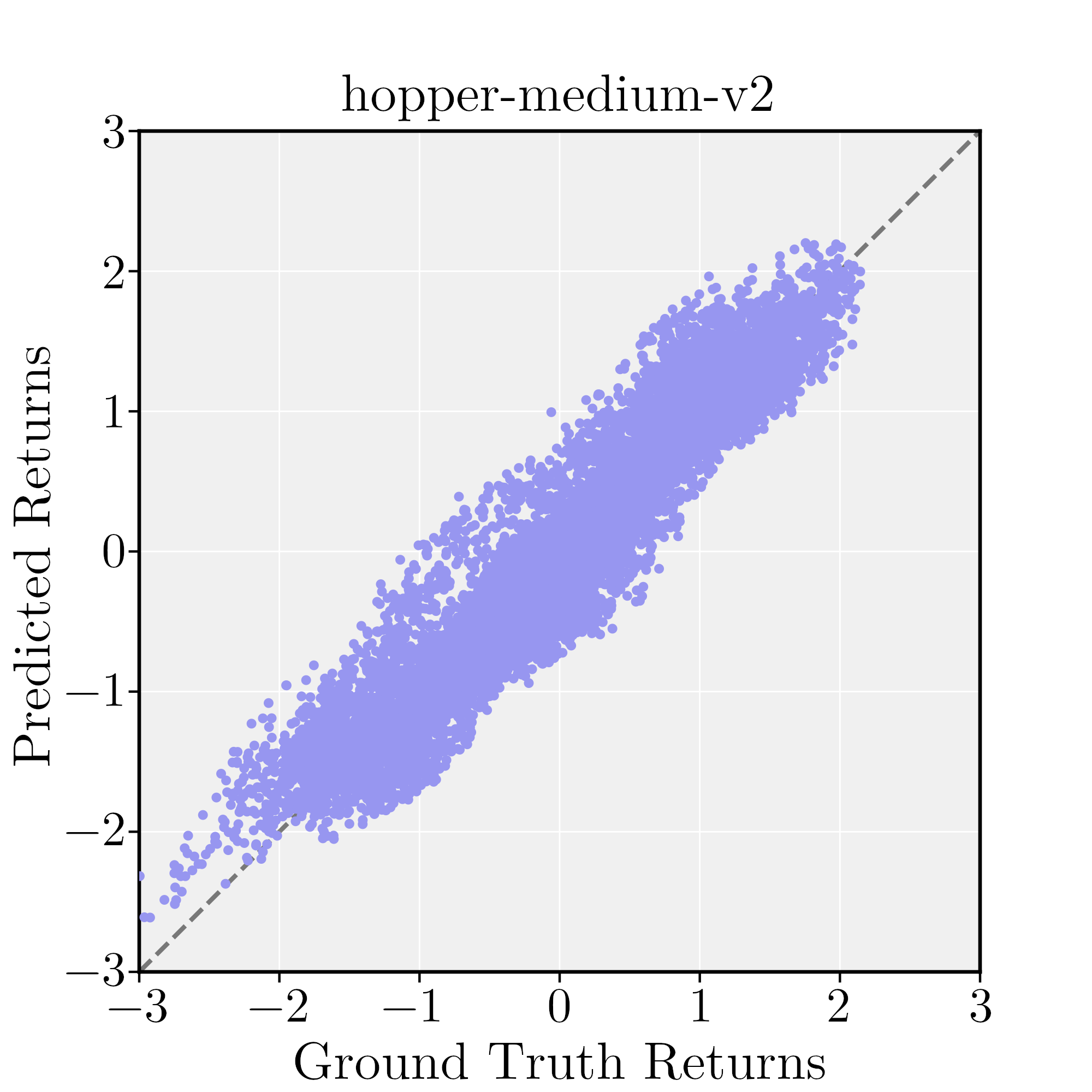}
    \end{minipage}
    \begin{minipage}[b]{0.3\textwidth}
            \centering
            \includegraphics[width=\textwidth]{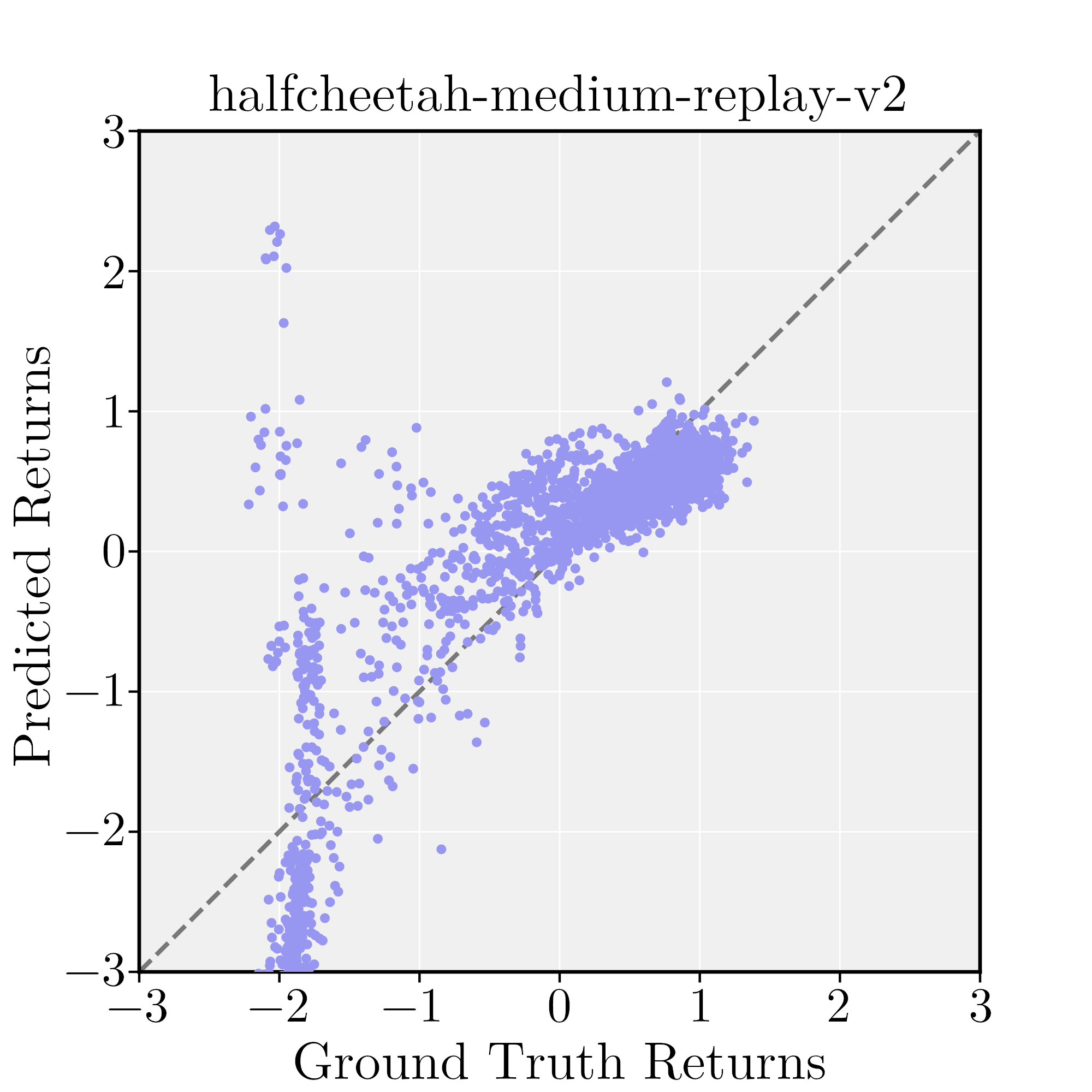}
    \end{minipage}
    \begin{minipage}[b]{0.3\textwidth}
            \centering
            \includegraphics[width=\textwidth]{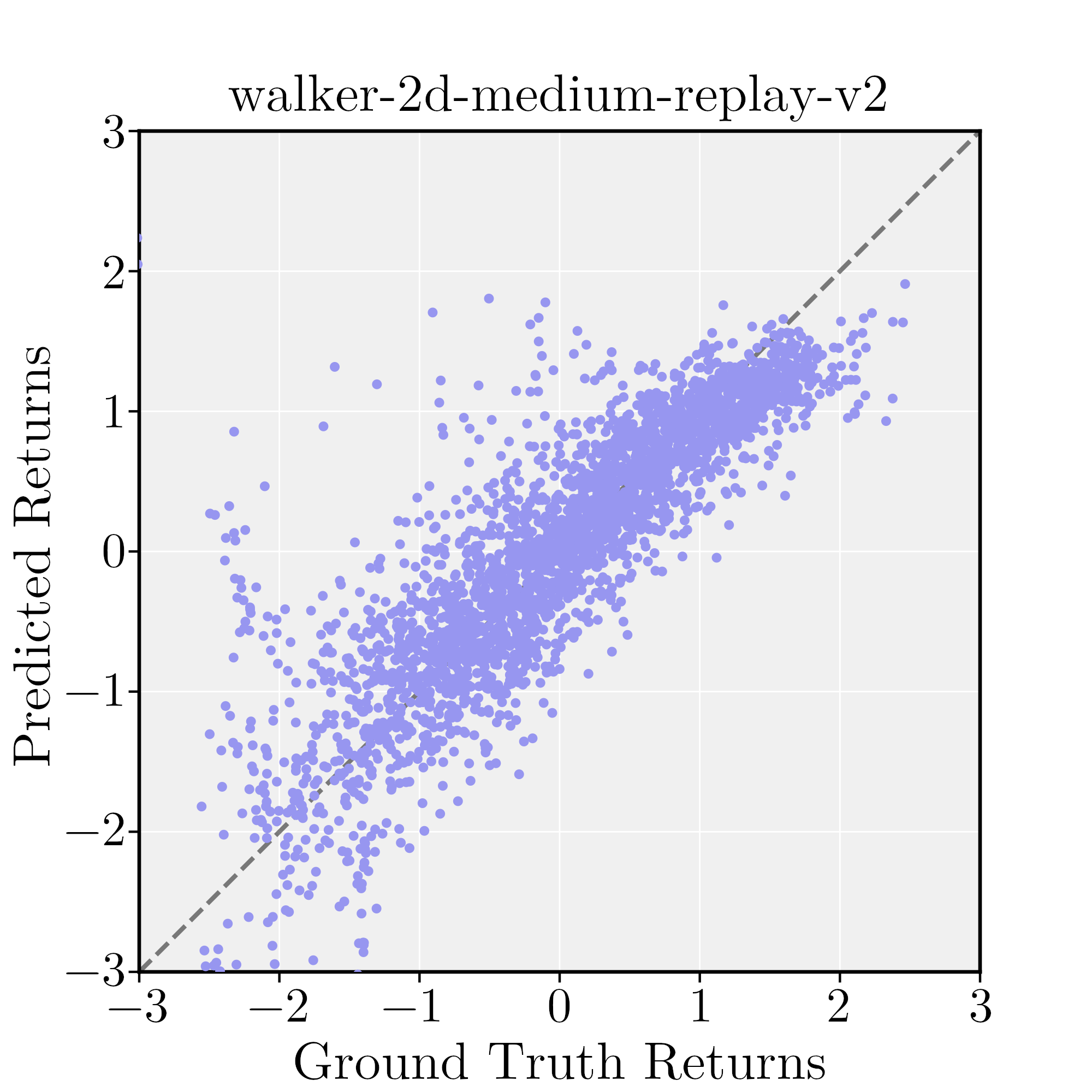}
    \end{minipage}
    \begin{minipage}[b]{0.3\textwidth}
            \centering
            \includegraphics[width=\textwidth]{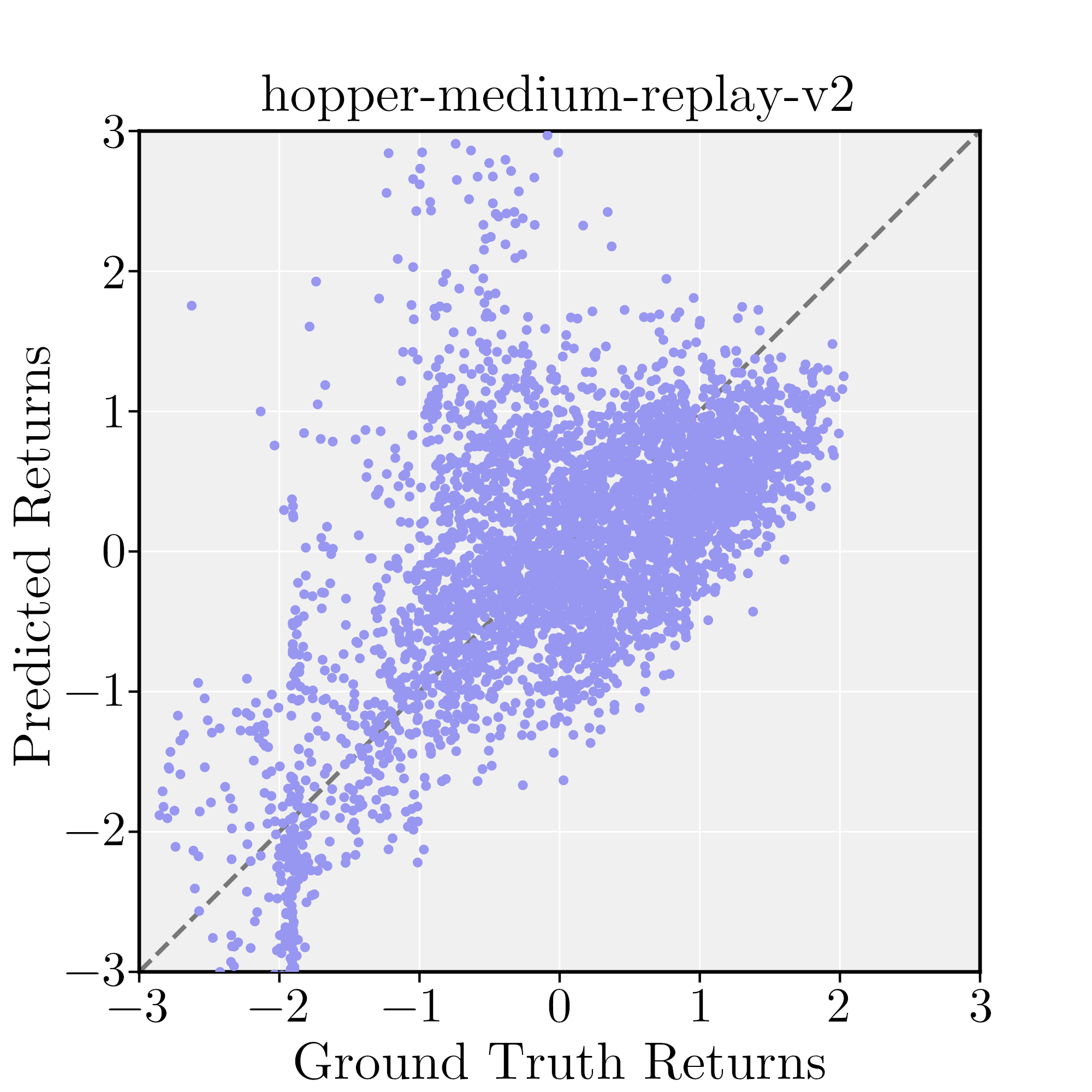}
    \end{minipage}
    \caption{Comparison of synergy-based/true rewards for different datasets in D4RL.
        The vertical axes correspond to the cumulative rewards over 200 steps estimated by SSRR, and the horizontal axes correspond to the true cumulative rewards over 200 steps.}
    \label{fig:reward_comparison}
\end{figure*}

\begin{table}[tb]
    \centering
    \caption{Correlation coefficients between the synergy-based and true rewards in D4RL}
    \begin{tabular}{lc}
        \toprule
        Dataset                      & Correlation coefficient \\
        \midrule
        halfcheetah-medium-expert-v2 & $0.9959$                \\
        walker2d-medium-expert-v2    & $0.8655$                \\
        hopper-medium-expert-v2      & $0.8718$                \\
        halfcheetah-medium-v2        & $0.9482$                \\
        walker2d-medium-v2           & $0.5809$                \\
        hopper-medium-v2             & $0.9526$                \\
        halfcheetah-medium-replay-v2 & $0.8039$                \\
        walker2d-medium-replay-v2    & $0.8263$                \\
        hopper-medium-replay-v2      & $0.5314$                \\
        \bottomrule
    \end{tabular}
    \label{tab:correlation_coefficients_ssrr_d4rl}
\end{table}

\begin{table*}[tb]
    \centering
    \caption{Performance of each algorithm in D4RL (Average sum of rewards over 10 episodes)}
    \begin{tabular}{lcc|c}
        \toprule
        Dataset                      & BC                   & SynIL (Ours) & IQL (ground-truth reward) \\
        \midrule
        halfcheetah-medium-v2        & $42.7 \pm 0.3$       & ${\bf 45.4 \pm 0.2}$   & $47.4 \pm 0.2$   \\
        hopper-medium-v2             & ${\bf 54.4 \pm 1.7}$ & $46.6 \pm 3.8$         & $54.4 \pm 3.5$   \\
        walker2d-medium-v2           & $71.7 \pm 6.3$       & ${\bf 72.0 \pm 5.4}$   & $80.0 \pm 4.9$   \\
        halfcheetah-medium-replay-v2 & $37.1 \pm 2.4$       & ${\bf 40.6 \pm 1.4}$   & $44.3 \pm 0.5$   \\
        hopper-medium-replay-v2      & $14.4 \pm 2.4$       & ${\bf 99.9 \pm 0.5}$   & $66.5 \pm 8.1$   \\
        walker2d-medium-replay-v2    & $15.5 \pm 6.4$       & ${\bf 74.5 \pm 3.0}$   & $69.5 \pm 7.9$   \\
        halfcheetah-medium-expert-v2 & $60.9 \pm 5.0$       & ${\bf 90.9 \pm 1.3}$   & $87.0 \pm 3.1$   \\
        hopper-medium-expert-v2      & $51.4 \pm 1.8$       & ${\bf 80.9 \pm 8.0}$   & $2.2 \pm 0.0$    \\
        walker2d-medium-expert-v2    & $80.0 \pm 5.8$       & ${\bf 108.9 \pm 0.3}$  & $110.7 \pm 0.5$  \\
        \bottomrule
    \end{tabular}
    \label{tab:d4rl_results}
\end{table*}

\begin{table}[tb]
    \centering
    \caption{Performance comparison in Robomimic (average success rate, 3 seeds $\times$ 50 rollouts)}
    \begin{tabular}{ccc|c}
        \toprule
        Dataset       & BC              & \begin{tabular}{@{}c@{}}SynIL\\(Ours)\end{tabular} & \begin{tabular}{@{}c@{}}IQL\\(sparse reward)\end{tabular} \\
        \midrule
        \begin{tabular}{@{}c@{}}
            Can\\
            (MH)
        \end{tabular} & $56.7 \pm 10.0$ & ${\bf 94.0 \pm 3.3}$ & $67.3 \pm 5.0$ \\
        \begin{tabular}{@{}c@{}}
            Square\\
            (MH)
        \end{tabular} & $24.0 \pm 4.3$  & ${\bf 49.3 \pm 8.2}$ & $16.7 \pm 5.2$ \\
        \bottomrule
    \end{tabular}
    \label{tab:robomimic_results}
\end{table}

\begin{table}[tb]
    \centering
    \caption{Execution time in Robomimic (seconds)}
    \begin{tabular}{ccc|c}
        \toprule
        Dataset       & BC              & \begin{tabular}{@{}c@{}}SynIL\\(Ours)\end{tabular} & \begin{tabular}{@{}c@{}}IQL\\(sparse reward)\end{tabular} \\
        \midrule
        \begin{tabular}{@{}c@{}}
            Can\\
            (MH)
        \end{tabular} & $2.1 \pm 0.2$  & ${\bf 1.3 \pm 0.0}$ & $2.2 \pm 0.2$ \\
        \begin{tabular}{@{}c@{}}
            Square\\
            (MH)
        \end{tabular} & $2.9 \pm 0.0$  & ${\bf 2.0 \pm 0.1}$ & $3.4 \pm 0.2$ \\
        \bottomrule
    \end{tabular}
    \label{tab:robomimic_time}
\end{table}

\begin{figure*}[tb]
    \centering
    \includegraphics[width=\textwidth]{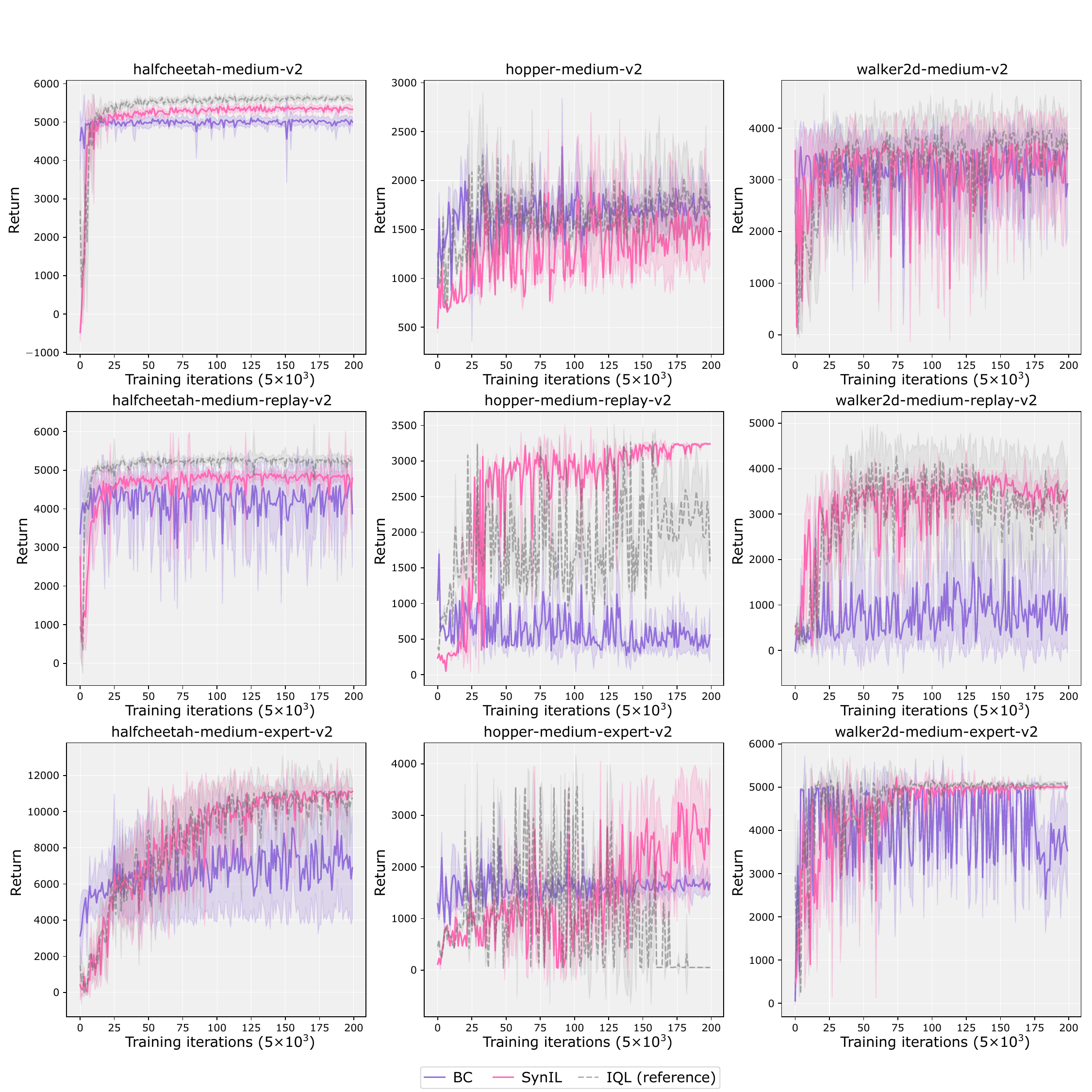}
    \caption{Learning curves of all environments.}
    \label{fig:lrcurve_d4rl}
\end{figure*}

\begin{figure*}[tb]
    \centering
    \includegraphics[width=0.5\textwidth]{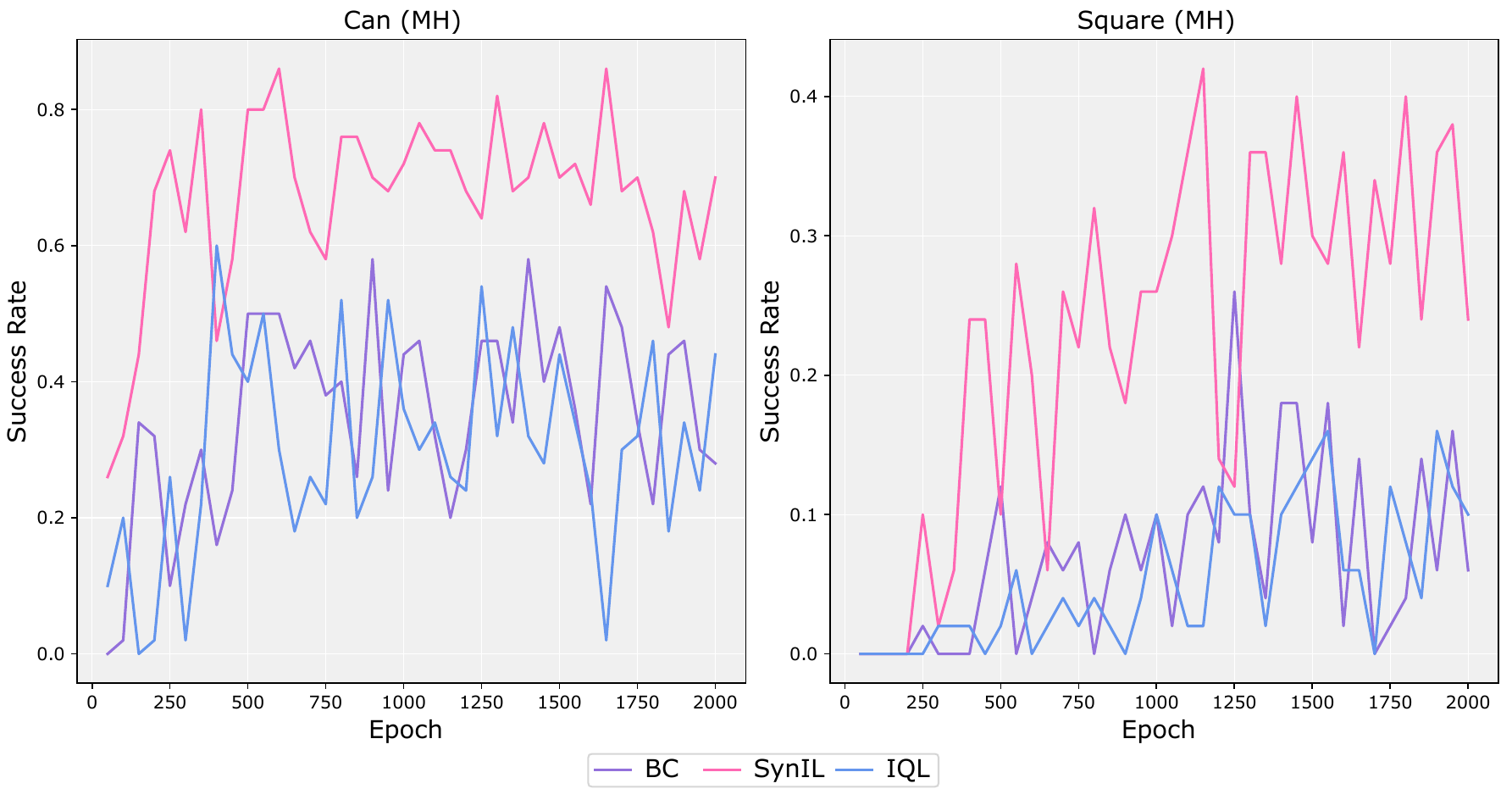}
    \caption{Learning curves of Robomimic.}
    \label{fig:lrcurve_robomimic}
\end{figure*}

\subsubsection{Correlation between Estimated Reward and Ground-Truth Reward}

First, we investigated the relationship between the synergy-based rewards and ground-truth rewards in D4RL datasets.
Figure~\ref{fig:reward_comparison} shows scatter plots of the rewards estimated by the SSRR (vertical axis, cumulative reward over 200 steps) and true rewards (horizontal axis, cumulative reward over 200 steps).
Additionally, Table~\ref{tab:correlation_coefficients_ssrr_d4rl} presents the correlation coefficients between the rewards estimated by the SSRR and true rewards.

As a result, the correlation coefficients were more than $0.8$ for all datasets except for the walker2d-medium-v2 and hopper-medium-replay-v2.
Particularly, the correlation coefficients were more than $0.86$ in all medium-expert datasets.

\subsubsection{Performance Comparison}

We evaluated the performance of imitation learning of SynIL (the proposed method), BC, and IQL in D4RL datasets.
Table~\ref{tab:d4rl_results} shows the performance, which was computed as the average sum of rewards over 10 episodes.
The values were normalized according to \citet{fu2020d4rl}.

As a result, according to Table~\ref{tab:d4rl_results}, SynIL outperformed BC on most datasets and was comparable to IQL with true rewards.
This is particularly evident in datasets of varying quality such as medium-replay and medium-expert.
On the other hand, BC, which performs imitation by supervised learning, shows decreased performance on datasets with diverse quality like medium-expert.

We then evaluated the performance of imitation learning in Robomimic environments.
Table~\ref{tab:robomimic_results} presents the performance, which was evaluated as the mean success rates for 50 trials across three different initial values (i.e. a total of 150 trials).
Table~\ref{tab:robomimic_time} shows the execution time in seconds.
Note that, since the Robomimic datasets did not contain reward data, we used sparse rewards, where a reward of $1$ was applied at the final steps of successful episodes, and $0$ was applied otherwise.

As a result, according to Table~\ref{tab:robomimic_results}, SynIL significantly outperformed both BC and even IQL.
Additionally, from Table~\ref{tab:robomimic_time}, it is evident that SynIL had a shorter execution time than both BC and IQL.
Learning curves in D4RL and Robomimic are shown in Figures~\ref{fig:lrcurve_d4rl} and \ref{fig:lrcurve_robomimic}.

\section{Discussion}

\subsection{Effectiveness of Synergy Rewards in Imitation Learning}

The results suggest that the synergy-based rewards, calculated from the degree of synergy manifestation, can be used as surrogate rewards quantifying the quality of demonstrations.
The synergy-based rewards have high correlation with the ground-truth rewards, as shown in Tables~\ref{tab:correlation_coefficients_ssrr_d4rl}.
Also, the use of the synergy-based rewards resulted in high performance in imitation learning, as shown in Tables~\ref{tab:d4rl_results}, and \ref{tab:robomimic_results}.
Those results demonstrate the effectiveness of the proposed method, that is, the use of synergy for evaluating the quality of demonstrations.

The improvement over BC is explained as selective use from mixed-quality data.
BC treats all collected actions as targets, suffering from suboptimal samples.
In contrast, SynIL first constructs a continuous quality signal, then allows IQL to propagate that signal through a long-horizon value function.
High-quality segments can contribute even if they occur within a mediocre trajectory, while coordinated failures can be downweighted if their inferred long-term value is low.
This argument is supported by the results of Hopper and Walker2d medium-replay and all medium-expert datasets, as described in Table~\ref{tab:d4rl_results}; these are precisely the settings where the dataset contains behavior from substantially different stages of learning.
Hopper medium is the exception, showing that the proxy can be unhelpful when quality differences are weak, when relevant task aspects are not expressed by linear coordination, or when reward-model errors alter IQL training.

The use of synergy-based rewards even outperformed IQL with ground-truth rewards in some cases of D4RL and in Robomimic.
This is because the ground-truth rewards we used in Robomimic were sparse, which was hard to train a well-performing policy.
The synergy-based rewards, in contrast, could densely provide more information and function as an important guide for successful completion in the task.
This also highlights the advantage of the synergy-based rewards that densely evaluate movement coordination.
Note that this comparison should not be interpreted as proof that synergy is a more correct objective than the environment reward.
Offline RL performance is sensitive to reward scale, terminal handling, implementation details, and hyperparameters.
The feasible conclusion is that the synergy-based rewards contain enough structure to support strong offline policy learning in these datasets.

\subsection{Relationship Between Motor Learning and Synergy}

The results shown in Tables~\ref{tab:d4rl_results} and \ref{tab:robomimic_results} suggest that the synergy-based reward was effective in improving the performance of imitation learning.
Here, although it has not fully been proven, we discuss why the synergy worked well in D4RL and Robomimic.

The success of the synergy-based method in D4RL would be caused by the nature of reinforcement learning.
The demonstraions in D4RL datasets have been generated by neural-network policies trained by deep reinforcement learning.
In reinforcement learning, an agent initially takes random state-action pairs, but as learning progresses, it takes more regular states and actions.
During this learning process, the dimensionality of the state-action distribution decreases.
This links to the increase of synergy manifestation.
Therefore, learning state-action relationship where synergy is highly manifested will result in the selective learning of more proficient behaviors.
This would be a reason why the proposed method resulted in a high performance in the D4RL datasets.
A similar phenomenon has been reported in \citet{chai2020motor}; the degree of synergy manifestation increases as deep reinforcement learning progresses.
In future work, these findings may lead to a new reinforcement learning framework that prefers to learn coordinated movements to accelerate motor learning.
Moreover, previous works have reported that synergies obtained in some tasks can generalize to new tasks \citep{AlBorno2020,kutsuzawa2022motor}.
These findings also suggest that the combination of synergy and reinforcement learning can improve the ability of motor learning.

On the other hand, the reason of the success of the synergy-based method in Robomimic would relate to the nature of the human motor learning.
It is known that many human movements become more regular, stable, and low-dimensional as learning progresses \citep{davids2012ecological}.
Even in unpredictable tasks, diverse actions are initially taken, but ultimately they converge to typical patterns of adaptive responses due to the principle of optimality \citep{braun2009learning}.
In such cases, it is considered that synergy in state actions also manifests.

In addition to the above, the effectiveness of synergy can also be considered from the perspective of computational neuroscience.
A recent theory in computational neuroscience argues that human's central nervous systems perform recognition and action generation so as to minimize the variational free energy \citep{friston2007free}.
Minimizing the variational free energy over time requires to decrease the amount of uncertainty, resulting in reduction of the entropy of the sensory signals \citep{karl2012free}.
Reduction in the entropy will bring about more coordinated distributions of the sensory signals and actions, resulting in the increase of the degree of synergy manifestation.
Therefore, it is possible that as a human becomes more proficient in a task, the variational free energy decreases more efficiently, thereby the synergy is highly manifested.

\subsection{Limitations}

The main limitation is that synergy is not universally equivalent to optimality.
A repetitive failure can be strongly coordinated, while a successful recovery can be high-dimensional.
SynIL is most plausible when inefficient demonstrations contain unstable corrections or inconsistent coupling and when proficient demonstrations use repeatable motor patterns.
Some studies have reported the relationship between synergy manifestation and optimality \citep{Todorov2002,chai2020motor}, but we could not provide solid evidence its generality.

Another limitation of this study is hyperparameter tuning.
Synergy score, defined in Equation~\ref{eq.synergy_score}, has several hyperparameters: $\eta$, $H$, and $S$.
We need to tune them appropriately to bring out the potential of the proposed method.
In particular, in the evaluations, we manually tuned $\eta$ to obtain high correlation with true rewards.
This means we implicitly used true reward information when tuning $\eta$, so the setup is not yet a fully reward-free procedure for hyperparameter selection.
However, even with this setup, it is possible to argue that synergy contains sufficient information to evaluate demonstration quality and construct a surrogate reward, given an appropriate parametrization.
We also used the same values across diverse tasks and agent morphologies in D4RL, suggesting that the hyperparameters are not very task-sensitive.

\section{Conclusion}

To overcome the vulnerability of imitation learning to imperfect demonstration datasets, this study introduced SynIL, an offline reinforcement learning framework utilizing synergy-derived quality evaluation.
We demonstrated that biological motor synergy, extracted algorithmically via PCA, serves as an effective quality score for unannotated trajectory segments and successfully generates dense rewards through self-supervised regression.
SynIL outperformed Behavior Cloning and demonstrated strong performance across benchmark locomotion and human teleoperation datasets, confirming the utility of synergy as a surrogate reward signal.
Beyond robotic control, our findings underscore the potential of integrating neuroscientific principles into machine learning pipelines.
Future work will aim to address current limitations, particularly by eliminating the reliance on ground-truth rewards for hyperparameter selection, clarifying the conditions under which synergy aligns with task optimality, and deploying the proposed framework to real-world physical AI platforms.

\printcredits

\bibliographystyle{cas-model2-names}

\bibliography{ref}



\end{document}